%% file: workshop_paper.tex
\documentclass[10pt,twocolumn,letterpaper]{article}

\usepackage{amssymb,amsmath}
\usepackage{epsfig}
\usepackage{rotating}
\usepackage{array}
\usepackage[ruled]{algorithm2e}
\usepackage{color}
\usepackage{multirow}
\usepackage{algorithmic}
\usepackage{subfigure}
\usepackage{gensymb}
\usepackage{tensor}
\usepackage{times}
\usepackage[pagebackref=false,breaklinks=true,letterpaper=true,colorlinks=
true,citecolor=black,bookmarks=false]{hyperref}

\usepackage{siunitx}
\date{\vspace{-5ex}}

\begin{document}

\title{Back to Basics:  Unsupervised Learning of Optical Flow via Brightness Constancy and Motion Smoothness}

\author{Jason J. Yu, Adam W. Harley and Konstantinos G. Derpanis\\
Department of Computer Science\\
Ryerson University, Toronto, Canada\\
{\tt\small \{jjyu, aharley, kosta\}@scs.ryerson.ca}
}

\maketitle


\input{abstract}

\section{Introduction}
\input{motivation}
\input{related_work}
\input{contributions}
\section{Technical approach}
\input{technical_approach}



\section{Empirical evaluation}
\input{evaluation}






\section{Discussion and summary}
\input{summary}


\small
\vspace{-5pt}\bibliography{bibref_definitions_short,bibref}
\bibliographystyle{plain}

\end{document}

%% file: abstract.tex
\begin{abstract}
\vspace{-7pt}Recently, convolutional networks (convnets) have proven useful for predicting optical flow.
Much of this success is predicated on the availability of large datasets that require expensive and involved
data acquisition and laborious labeling.  To bypass these challenges,
we propose an unsupervised approach (i.e., without 
leveraging groundtruth flow) to train a convnet end-to-end for predicting optical flow between two images.
We use a loss function that combines a data term that measures photometric constancy over time with a spatial
term that models the expected variation of flow across the image. 
Together these losses form a proxy measure for losses based on the groundtruth flow.
Empirically, we show that a strong convnet baseline trained with the proposed unsupervised approach outperforms
the same network trained with supervision on the KITTI dataset. 
\end{abstract}

%% file: motivation.tex
Visual motion estimation is a core research area of computer vision.  Most prominent has been the recovery of the
apparent motion of image brightness patterns, i.e., optical flow.  Much of this work has centred on extracting the pixelwise
velocities between two temporal images within a variational framework \cite{horn1981,sun2014}.

Recently, convolutional networks (convnets) have proven useful for a variety of per-pixel prediction tasks, including optical flow \cite{fischer2015}.
Convnets are high-capacity models that approximate the complex, non-linear transformation between input imagery and the output.  
Success with convnets has relied almost exclusively on fully-supervised schemes, where the target value
(i.e., the label) is provided during training.  This is problematic for learning optical flow because directly obtaining the motion field
groundtruth from real scenes {}\textemdash{} the quantity that optical flow attempts to approximate {}\textemdash{} is not possible.  

In this paper, we propose an end-to-end unsupervised approach to train a convnet for predicting optical flow between two images 
based on a standard variational loss.
Rather than rely on imagery as well as the corresponding groundtruth flow for training, we use the images alone.  In particular,
we use a loss function that combines a data term that measures photometric constancy over time with a spatial
term that models the expected variation of flow across the image. 
The photometric loss measures the difference between the first input image and the (inverse) warped subsequent image based on the predicted
optical flow by the network.  The smoothness loss measures the difference between spatially neighbouring flow predictions.  Together, these
two losses form a proxy for losses based on the groundtruth flow.

%% file: related_work.tex
Recovering optical flow between two frames is a well studied problem, with much previous work founded on variational formulations \cite{horn1981,brox2011,sun2014,revaud2015}. 
Our loss is similar to the objective functions proposed for two-frame motion estimation; however, rather than 
optimize the velocity map between input frames,
we use it to optimize the convnet weights over the training set of imagery. 

Several recent works \cite{fischer2015,mayer2016} 
have proposed convnets that learn the 
mapping between input image frames and the corresponding flow.
Each of these approaches is presented in a supervised setting, where images and their corresponding groundtruth flows are provided.
This setting assumes the availability of a large, annotated dataset. 
Existing flow datasets (e.g., KITTI  \cite{geiger2013}) 
are too small to support training accurate networks.  Computer generated scenes and their corresponding flow \cite{fischer2015,mayer2016,gaidon2016} 
provide a means to address this issue.
Although some recent efforts have attempted to semi-automate the data creation process \cite{gaidon2016}, creating large, diverse imagery remains laborious.
 Another possibility is using the output of an existing optical
flow estimator to provide the groundtruth \cite{tran2016}.  This training approach may result in learning both correct flow prediction and the failure aspects of the flow estimator
used for training.  In this work, we avoid these drawbacks by learning flow in an unsupervised manner, using only  the input imagery. 

Concurrent work has proposed unsupervised methods to circumvent the need of vasts amounts of labeled data for training. 
A spatiotemporal video autoencoder \cite{patraucean2015} was introduced that incorporates a long short-term memory (LSTM) architecture for unsupervised
flow and image frame prediction.  
Here, we present a simpler feedforward convnet model targeting flow prediction alone.
Most closely related to the current paper is recent work that proposed a convnet for depth estimation trained in an unsupervised manner \cite{garg2016}.  In a similar fashion to the proposed approach, a  photometric loss warps one image to another, and a smoothness loss term is used to bias the predictor towards smooth depth estimates.
Unlike the current work, the manner in which the photometric loss is handled (via a linear Taylor series approximation) precludes end-to-end learning.

%% file: contributions.tex
\vspace{5pt}\noindent{\bf Contributions } In the light of previous research, we make the following contributions.  First, we present an unsupervised approach to training a convnet in an
end-to-end manner for predicting optical flow between two images.  The limited but valuable  
groundtruth flow is reserved for fine-tuning the network and cross-validating its parameters.
Second, we demonstrate empirically that a strong convnet baseline trained with our unsupervised approach outperforms the same network trained with supervision on KITTI,
where insufficient groundtruth flow is available for training.

%% file: technical_approach.tex
Given an RGB image pair as input, $\mathbf{X} \in \mathbb{R}^{H\times W \times 6}$, our objective is to learn a non-linear mapping (approximated by a convnet) 
to the corresponding optical flow, $\mathbf{Y} \in \mathbb{R}^{H\times W \times 2}$, where $H$ and $W$ denote the image height and width, respectively.
In Section \ref{sec:loss}, we outline our unsupervised loss.  Section \ref{sec:architecture} provides details on how the unsupervised loss is integrated with a 
reference convnet architecture.

\subsection{Unsupervised loss} \label{sec:loss}

The training set is comprised of pairs of temporally consecutive images, $\{ I(x,y,t), I(x,y,t+1)\}$.  Unlike prior work, we do not assume access to the corresponding velocity pixel labels, cf.\ \cite{fischer2015}.
Instead, we return to traditional means for scoring a given solution, via a loss that combines a photometric loss 
between the first image and the warped second image, and a loss related to the smoothness of the velocity field prediction \cite{horn1981}:
\begin{align}
 \mathcal{L}&(\mathbf{u}, \mathbf{v}; I(x,y,t), I(x,y,t+1)) = \nonumber\\
  &\ell_\text{photometric} (\mathbf{u}, \mathbf{v}; I(x,y,t), I(x,y,t+1)) + \nonumber\\
  &\lambda \ell_\text{smoothness} (\mathbf{u}, \mathbf{v}), \label{eq:loss}
\end{align}
where 
$\mathbf{u}, \mathbf{v} \in \mathbb{R}^{H\times W}$ are the horizontal and vertical components of the predicted flow field, respectively, and $\lambda$ is a regularization parameter
that weighs the relative importance of smoothness of the predicted flow.  Note, the photometric loss can be replaced or augmented with other measures, 
such as the image gradient constancy \cite{brox2011}. 

Given the predicted flow, the photometric loss is computed as the difference between the first image and the backward/inverse warped second image:
\begin{align}
    \ell&_\text{photometric} (\mathbf{u}, \mathbf{v}; I(x,y,t), I(x,y,t+1)) = \\\nonumber &\sum_{i,j} \rho_D(I(i,j,t) - I(i + u_{i,j}, j + v_{i,j}, t+1)),
\end{align}
where $\rho_D$ is the data penalty function.
We consider the robust generalized Charbonnier penalty function $\rho(x) = (x^2 + \epsilon^2)^\alpha$ to mitigate the effects of outliers \cite{sun2014}.

To compute the non-rigid backward warp, we use the recently proposed spatial transformer module \cite{jaderberg2015}. 
This allows the learning to be performed with standard backpropagation
in an end-to-end fashion.
In brief, the spatial transformer can be described as two parts that work in sequence: (i) a sampling grid generator and (ii) a differentiable image sampler. (The spatial 
transformer localization step is not needed here as flow prediction, $(u,v)$, provides the necessary parameters for the mapping between image points across frames.)  The sampling grid is generated by
the following pointwise transformation:
\begin{equation}
   \begin{pmatrix}
      x_2 \\ y_2   
      \end{pmatrix} = 
      W_{(u, v)}  \begin{pmatrix} x_1 \\ y_1   \end{pmatrix} =  \begin{pmatrix} x_1 + u \\ y_1 + v   \end{pmatrix},
\end{equation}
where $(x_1, y_1)$ are the coordinates in the first image and $(x_2, y_2)$ are the sampling coordinates in the second image. 
The bilinear sampling step can be written in the following \mbox{(sub-)differentiable} form:
\begin{align}
   I&_\text{warp}(x_1,y_1,t+1) = \nonumber\\ &\sum_j^H\sum_i^WI(i, j, t+1)M(1 - |x_2-i|)M(1 - |y_2-j|),
\end{align}
where $M(\cdot) = \max(0, \cdot)$. 
For details about backpropagating through this module, see \cite{jaderberg2015}.

Regions with insufficient image structure support 
multiple equally scoring velocities, e.g., the aperture problem.  To address this ambiguity, we introduce a standard robust (piecewise) smoothness loss: 
\begin{align}
 \ell&_\text{smoothness} (\mathbf{u}, \mathbf{v}) =\nonumber\\
   &  \sum_j^H\sum_i^W [\rho_{S} (u_{i,j} - u_{i+1,j})  + \rho_{S} (u_{i,j} - u_{i,j+1}) \nonumber\\  
   & + \rho_{S} (v_{i,j} - v_{i+1,j}) + \rho_{S} (v_{i,j} - v_{i,j+1})],
   \end{align}
where $\rho_S(\cdot)$ is the (spatial) smoothness penalty function realized by
the generalized Charbonnier function. 

\begin{figure}[t]
\begin{center}
    \epsfig{file=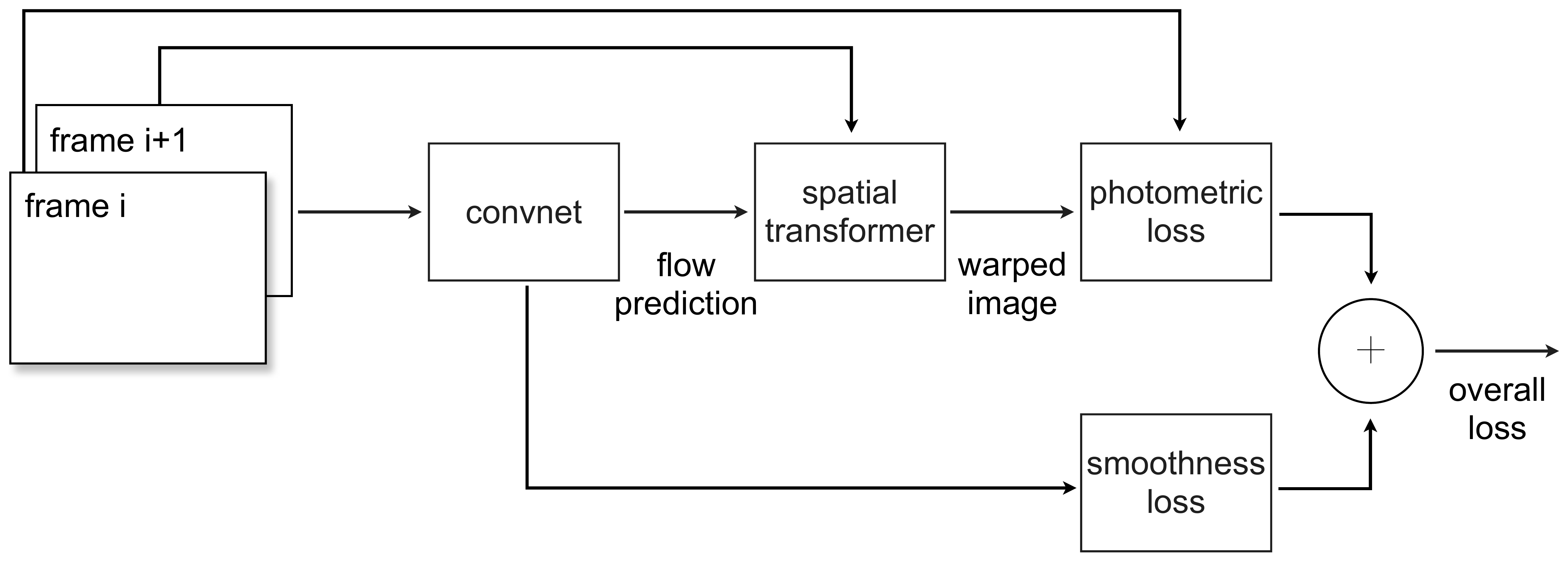, width = 3.2in}
\end{center}
\vspace{-20pt}
\caption{Overview of our unsupervised  approach.}
\label{fig:overview}
\vspace{-10pt}
\end{figure}

A summary of our proposed unsupervised approach for flow prediction is provided in Fig.\ \ref{fig:overview}.

\subsection{Network architecture}  \label{sec:architecture}
\begin{figure}[t]
\begin{center}
    \epsfig{file=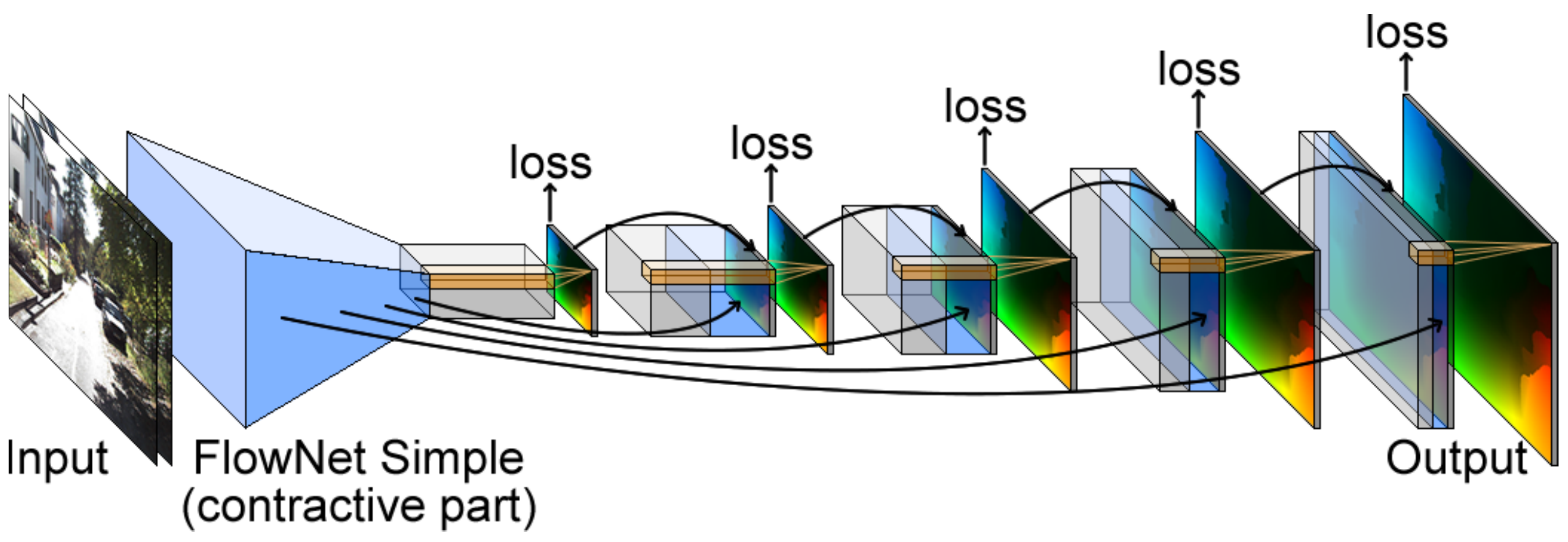, width =2.9in}
    \end{center}
 \vspace{-20pt}
\caption{``FlowNet Simple'' architecture. Two images are taken as input, and an optical flow prediction is generated using a multi-stage refinement process.
Feature maps from the contractive part, as well as intermediate flow predictions, are used in the ``upconvolutional'' part.}
\label{fig:flownet_architecture}
\vspace{-10pt}
\end{figure}

We use ``FlowNet Simple'' \cite{fischer2015} as a reference network.
This architecture consists of a 
contractive part followed by an expanding part.  
The contractive part takes as
input two RGB images stacked together, and processes them with a cascade of strided convolution layers.
The expanding part implements a ``skip-layer'' architecture that combines information from various levels of the contractive part with ``upconvolving'' layers to iteratively refine the coarse flow predictions.  The FlowNet Simple architecture is illustrated in Fig.\ \ref{fig:flownet_architecture}.

In this work, we use a loss comprised of a final  loss and several intermediate losses placed at various stages of the expansionary part.  The intermediate losses 
are meant to guide earlier layers more directly towards the final objective \cite{lee2015}.  
In FlowNet, the endpoint error (EPE), a standard error measure for optical flow, is used as the supervised training loss.  As a proxy
to per-pixel groundtruth flow, we replace the EPE with the proposed unsupervised loss, (\ref{eq:loss}).

%% file: evaluation.tex
\subsection{Datasets}

 
\noindent{\bf Flying Chairs } This synthetic dataset is realized by applying affine mappings to publicly available
colour images and a rendered set of 3D chair models.  The dataset contains $22, 232$ training 
and 640 test image pairs with groundtruth flow.  
To cross-validate the hyper-parameters and monitor for
overfitting in learning, we set aside $2, 000$ image pairs from the training set.  
We use both photometric and geometric augmentations to avoid overfitting. 
The photometric augmentations are comprised of additive Gaussian noise applied to each image,
contrast, multiplicative colour changes to the RGB channels, gamma  and additive
brightness.  
The geometric transformations are comprised of 2D translations, left-right flipping, 
rotations and scalings.\\ 

\noindent{\bf KITTI 2012 }  This dataset consists of images collected on a driving platform.
There are 194 and 195 training and testing image pairs, respectively, with sparse groundtruth flow.
The training set is used for cross-validation and to monitor the learning progress.
For training, we use the raw KITTI data 
from the city, residential and road classes, where groundtruth flow is unavailable.  
To avoid training on related testing imagery, we remove all raw images that are visually similar with the testing ones, including their 
temporal neighbours $\pm 20$ frames.  The (curated) raw data is comprised of $82, 958$  image pairs.
We include both photometric and geometric augmentations.
We use the same type of photometric augmentations as applied to Flying Chairs. 
The geometric transformations consist of left-right flipping and scalings. 
We also use a small relative translation.

\subsection{Training details}
We use the ``FlowNet Simple'' architecture provided in the publicly available FlowNet Caffe code \cite{FlowNet}. 
For the photometric loss, the generalized Charbonnier parameter, $\alpha$, is set to $0.25$ and  $0.38$ for Flying Chairs and KITTI, respectively.
For the smoothness loss, $\alpha$ is set to $0.37$ and $0.21$ for Flying Chairs and KITTI, respectively.
The smoothness weight, $\lambda$, is set to $1$ for Flying Chairs and $0.53$ for KITTI. 
We use Adam as the optimization method, where its parameters $\beta_1=0.9$ and $\beta_2=0.999$.  
The initial learning rate is set to 
\num{1.6e-5} 
for Flying Chairs and 
\num{1.0e-5} 
for KITTI and 
we divide it by half every $100,000$ iterations.
The batch size is set to four image pairs.  In total, we train using $600, 000$ iterations for Flying Chairs and $400, 000$ for KITTI.
In initial tests, we noticed that the unsupervised approach had difficulties in regions that were highly saturated or very dark. 
Adding photometric augmentation compounds this issue by making these regions even less discriminable.  To address this issue, we pass
the geometrically augmented images directly to the photometric loss prior to photometric augmentation.  Further,
we apply a local $9\times 9$ response normalization to the geometrically augmented images to ameliorate multiplicative lighting factors.


\subsection{Results}

\begin{figure}[t]
\begin{center}
    \epsfig{file=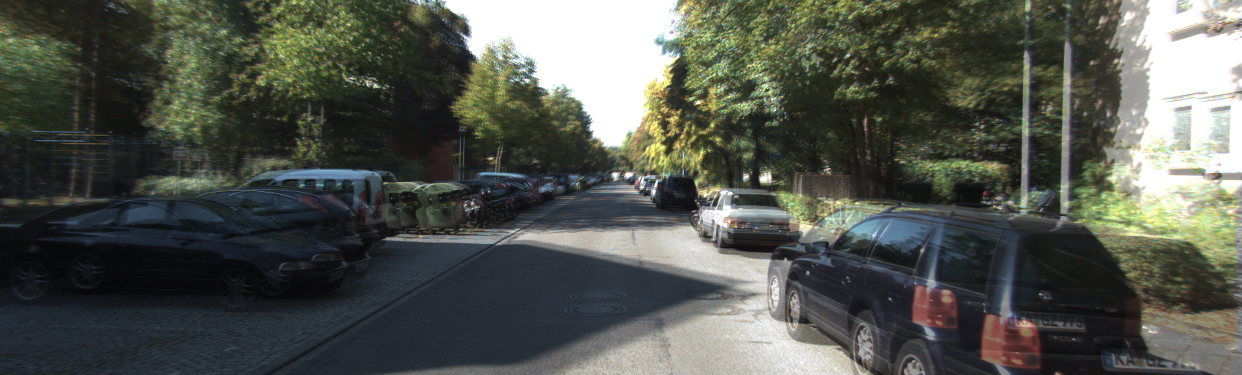, width = 2.9in}
    \epsfig{file=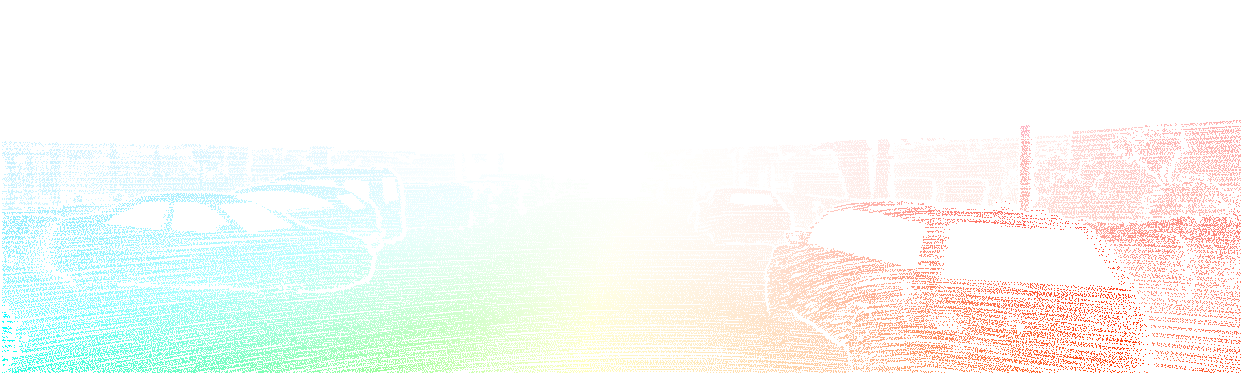, width = 2.9in}
    \epsfig{file=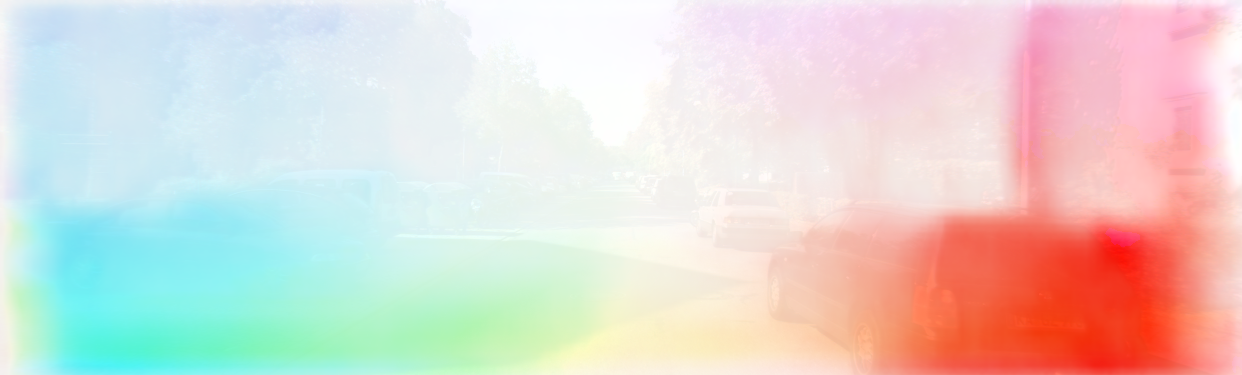, width = 2.9in}
\end{center}
\vspace{-20pt}
\caption{KITTI  example. 
(top-to-bottom) Input image frames overlaid, groundtruth flow, and predicted flow from unsupervised network overlaid on the first input image.} 
\label{fig:flow-example}
\vspace{-5pt}
\end{figure}

Table \ref{table:EPE} provides a summary of results.  As expected,  FlowNet trained with the groundtruth flow on Flying Chairs outperforms the unsupervised one.  Note, however,  the scenario where sufficient dense groundtruth is available is generally unrealistic with real imagery. 
Conversely, KITTI  exemplifies an automative scenario, where abundant dense groundtruth flow with real images is unavailable.  To sidestep this issue, previous work \cite{fischer2015} 
used synthetic data as a proxy for supervised training.
On the non-occluded (NOC) metric, the unsupervised approach improves upon the 
supervised one \cite{fischer2015} on the KITTI training set.  This improvement persists on the official test set.  Considering all pixels (i.e., occluded and non-occluded)
the proposed approach remains competitive to the supervised one.  Figure \ref{fig:flow-example} shows an example flow prediction result on KITTI.
While the performance of the unsupervised approach lags behind the state-of-the-art, it operates in realtime with a testing runtime of 
$0.03$ seconds on an NVIDIA GTX 1080 GPU.




\begin{table}[t]\begin{center}
\begin{tabular}{|c|c|c|c|c|c|}\hline
        \multirow{3}{*}{Approach}       & \multirow{2}{*}{Chairs} & \multicolumn{4}{c|}{KITTI} \\\cline{3-6}
                        &   &\multicolumn{2}{c|}{Avg.\ All}     &  \multicolumn{2}{c|}{Avg.\ NOC}\\\cline{2-6}
                        &test  &train & test&train & test\\\hline\hline
   EpicFlow \cite{revaud2015}    &2.9& 3.5& 3.8& 1.8& 1.5  \\
   DeepFlow \cite{weinzaepfel2013}  &3.5& 4.6& 5.8& 2.0& 1.5   \\
   LDOF \cite{brox2011}        &3.5& 13.7& 12.4& 5.0& 5.6 \\
   FlowNet \cite{fischer2015}     &2.7&7.5 &9.1& 5.3 &5.0    \\\hline
   FlowNet (ours)    & 5.3 & 11.3 & 9.9 & 4.3 & 4.6 \\ \hline   
\end{tabular}
\end{center}
\vspace{-17pt}
\caption{Average endpoint error (EPE) flow results.  
The reported EPEs for supervised FlowNet are the best results of the FlowNet Simple architecture \cite{fischer2015}
without variational smoothing post-processing.
``Avg.\ All'' and ``Avg.\ NOC'' refer to the EPE taken over all the labeled pixels and all non-occluded labeled pixels, respectively.}
\label{table:EPE}
\vspace{-10pt}
\end{table}

%% file: summary.tex
We presented an end-to-end unsupervised approach to training convnets for optical flow prediction. 
We showed that the proposed unsupervised training approach yields competitive and even superior performance
to a supervised one. 
This opens up avenues for further improvement by 
leveraging the vast amounts of video that can easily be captured with commodity cameras taken in the domain of interest, such as automotive applications.
Furthermore, this is a general learning framework that can be extended in a variety of ways via more sophisticated losses
to enhance convnet-based mappings between temporal input imagery and flow.
